\documentclass[a4paper]{article}
\usepackage{epsfig}

\def\bbbr{{\rm I\!R}} 
\def\bbbz{{\mathchoice {\hbox{$\sf\textstyle Z\kern-0.4em Z$}}
{\hbox{$\sf\textstyle Z\kern-0.4em Z$}}
{\hbox{$\sf\scriptstyle Z\kern-0.3em Z$}}
{\hbox{$\sf\scriptscriptstyle Z\kern-0.2em Z$}}}}

\newcommand{\comment}[1]{
  {\hfill\raggedleft{
      \scriptsize{\textsl{\textsf{{\setlength{\baselineskip}{-100pt}%
        #1\\\mbox{}\vspace{-8pt}\mbox{}}}}}}}}

\newenvironment{program}{%
  \begin{footnotesize}
    \begin{description}%
      \setlength{\leftmargin}{70mm}%
      }%
    {\end{description} \end{footnotesize} \vspace{-0pt}}

\newcommand{\rand}{{\mathit{r}}}   
\newcommand{\xea}{\bar{x}^{E\hspace{-.6mm}A}}
\newcommand{\xeai}{\bar{x}^{{E\hspace{-.6mm}A}^i}}

\newcommand{\xeaf}{\bar{x}^{{E\hspace{-.6mm}A}^f}}
\newcommand{\xeam}{\bar{x}^{{E\hspace{-.6mm}A}^m}}
\newcommand{\zea} {\bar{z}^{{E\hspace{-.6mm}A}}}
\newcommand{\zeai}{\bar{z}^{{E\hspace{-.6mm}A}^i}}
\newcommand{\zeaj}{\bar{z}^{{E\hspace{-.6mm}A}^j}}
\newcommand{\zu}{\bar{z}^{\mathrm{up}}}
\newcommand{\zd}{\bar{z}^{\mathrm{down}}}
\newcommand{\xls}{\bar{x}^{{L\hspace{-.6mm}S}}}
\newcommand{\zls}{\bar{z}^{{L\hspace{-.6mm}S}}}
\newcommand{\yea}{{\bar{y}}}             
\newcommand{\xlp}{x^{L\hspace{-.6mm}P}}

\newcommand{\mcc}[1]{\multicolumn{1}{|c|}{#1}}

\newcommand{\mccr}[1]{\multicolumn{1}{|c||}{#1}}

\begin{document}

\title{An evolutionary solver for linear integer programming}

\author{Jo\~ao Pedro Pedroso\\
  Riken Brain Science Institute \\
  Hirosawa 2-1, Wako-shi, Saitama 351-0198, Japan\\
  e-mail: jpp@brain.riken.go.jp
  }              

\date{
  \begin{large}
    \textbf{BSIS Technical Report No.98-7}\\
  \end{large}
  August 1998
  }

\maketitle

\begin{abstract}
  In this paper we introduce an evolutionary algorithm for the
  solution of linear integer programs.  The strategy is based on the
  separation of the variables into the integer subset and the
  continuous subset; the integer variables are fixed by the
  evolutionary system, and the continuous ones are determined in
  function of them, by a linear program solver.
  
  We report results obtained for some standard benchmark problems, and
  compare them with those obtained by branch-and-bound.  The
  performance of the evolutionary algorithm is promising.  Good
  feasible solutions were generally obtained, and in some of the
  difficult benchmark tests it outperformed branch-and-bound.
\end{abstract}

\section{Introduction}
\label{sec:introduction}

Integer linear programming problems are widely described in the
combinatorial optimisation literature, and include many well-known and
important applications. Typical problems of this type include lot
sizing, scheduling, facility location, vehicle routing, and more; see
for example \cite{IntCombOptim,lscopt}.  The problem consists of
optimising a linear function subject to a set of linear constraints,
in the presence of integer and, possibly, continuous variables.  If
the subset of continuous variables is empty, the problem is called
\emph{pure integer} (IP).  In the more general case, where there are
also continuous variables, the problem is usually called \emph{mixed
  integer} (MIP).

The general formulation of a mixed integer linear program is
\begin{equation}
  \label{eq:mip}
  \max_{x,y}\{c x + h y : A x + G y \leq b,
  x \in \bbbz_+^n, y \in \bbbr_+^p\}
\end{equation}
where $\bbbz_+^n$ is the set of nonnegative integral $n$-dimensional
vectors and $\bbbr_+^p$ is the set of nonnegative $p$-dimensional
vectors.  $A$ and $G$ are $m \times n$ and $m \times p$ matrices,
respectively, where $m$ is the number of constraints.  The integer
variables are $x$, and the continuous variables are
$y$.

\subsection{The evolutionary structure}

The main idea for the conception of the algorithm described here is
that if the integer variables of a MIP are fixed, what remains to
solve is a standard LP problem; this can be done exactly and
efficiently, for example by the simplex algorithm or by interior point
methods.  We are therefore able to make the integer variables evolve
through an evolutionary algorithm (EA); after they are fixed by the
EA, we can determine the continuous variables in function of them.

\subsection{Branch-and-bound.}
The most well known algorithm for solving MIPs is branch-and-bound
(B\&B) (for a detailed description see, for
example,~\cite{IntCombOptim}).  This algorithm starts with a
continuous relaxation of the MIP, and proceeds with a systematic
division of the domain of the relaxed problem, until the optimal
solution is found.  There are two main advantages of the B\&B
algorithm.  The first and most important is that its solution is
optimal (or there are no feasible solutions); the other is that some
nodes of the B\&B exploration graph can be pruned, and therefore the
algorithm's speed and memory requirement improved.  These are two
important reasons to dissuade the application of an EA for the same
purpose: EAs cannot prove that the solution found is optimal, and in
what concerns convergence the best that can be proved is that, for
elitist EAs, we obtain a sequence of evaluations which converges to
the optimal objective value as the number of generations tends to
infinity.

We believe that it is nevertheless worthy to try to use an EA for this
type of problems, because of two other important reasons.  The first
is that it is easy to incorporate in the EA a problem-specific local
search method, possibly working on primal solutions, taking advantage
of the problem structure; this could provide a speedup of one order of
magnitude.  The second reason is that in some cases B\&B fails to find
a good feasible solution, sufficient for most practical applications,
in a reasonable computational time.  It can be hoped that an EA does
better than B\&B in these cases.

\subsection{Benchmark problems}
Instances of integer linear problems correspond to specifications of
the data: the matrices $A$ and $G$, and the vectors $b$, $c$ and $h$
in equation~\ref{eq:mip}.  The most commonly used representation of
instances of these problems is through \emph{MPS files}.  The format
of these files has the advantage of being standard, and hence readable
by most of the solvers; the disadvantage being that it can not provide
information concerning the specific characteristics of the problem.

We have tested the EA with a subset of the benchmark problems that are
available in the \emph{MIPLIB}~\cite{miplib}.  These problems range
from the moderately easy to the very difficult, for the solution
techniques available nowadays.

\section{The evolutionary operators}
\label{sec:evol-op}

Evolutionary algorithms function by maintaining a set of solutions,
generally called a \emph{population}, and making these solutions
evolve through operations that mimic the natural evolution:
reproduction, and selection of the fittest.  Some of these operators
where customised for the concrete type of problems that we are
dealing with; we focus on each of them in the following
sections.

\subsection{Representation of the solutions}
\label{sec:representation}

The part of the solution that is determined by the EA is the subset of
integer variables, $x$ in equation~\ref{eq:mip}.  Integer variables
are fixed by the EA, leading to an LP with only the continuous
variables, $y$, free; these are determined afterwards by solving a
linear problem.

We use the term \emph{individual} to mean a solution of the original
mixed-integer problem, and the term \emph{genome} to mean the subset
of integer variables of that solution.  The solution corresponding to
a particular individual is represented in the EA by its genome, an
$n$-dimensional vector $\xea = (\xea_1 \ldots \xea_n)$; we call each
$\xea_k$ a \textit{chromosome}.

An individual $i$ kept in the algorithm's population is hence
represented by the vector of integer variables $\xeai$, and the
corresponding vector of continuous variables $\yea^i$ is determined by
an LP solver, at the time of its evaluation.


\subsection{Evaluation of individuals}
\label{sec:objective}

The solutions that are kept by the algorithm---or, in other words, the
individuals that compose the population---may be feasible or not.  For
the algorithm to function appropriately it has to be able to deal with
both feasible and infeasible individuals coexisting in the population.

In the process of evaluation of an individual, we first formulate an
LP by fixing all the variables of the MIP at the values of the
individual's genome:
\begin{equation}
  \label{eq:lp-conv}
  z = \max_{y}\{c \xea + h y : G y \leq b - A \xea, y \in \bbbr_+^p\}
\end{equation}

We are now able to solve this (purely continuous) linear problem using
a standard algorithm, like the simplex.

\subsubsection{Feasible solutions}

If problem~\ref{eq:lp-conv} is feasible, the evaluation (fitness)
attributed to the corresponding individual is the objective value $z$,
and the individual is labelled feasible.  We denote this fitness by
$\zea$, a data structure consisting of the objective value and a
flag stating that the solution is feasible.

\subsubsection{Infeasible solutions}

If problem~\ref{eq:lp-conv} is infeasible, we formulate another LP for
the minimisation of the infeasibilities.  This is accomplished by
setting up artificial variables and minimising their sum (a procedure
that is identical to the phase I of the simplex algorithm):
\begin{equation}
  \label{eq:lp-infeas}
  \zeta = \min_{s}\{\sum_{k=1}^{m} s_k : G y \leq b - A \xea + s,
    y \in \bbbr_+^p \;,\;  s \in \bbbr_+^m\}
\end{equation}
where $m$ is the number of constraints.

The evaluation attributed to such an individual is the value~$\zeta$
of the optimal objective of the LP of equation~\ref{eq:lp-infeas}, and
the individual is labelled infeasible.  The fitness data structure
$\zea$ consists of the value~$\zeta$ and an infeasibility
flag.

\subsubsection{Comparison and selection of individuals}
\label{sec:comparison}

For the selection of individuals, we have to provide a way for
comparing them, independently of the corresponding solutions being
feasible or not.  What we propose is to rank the solutions, so that:
feasible solutions are always better than infeasible ones, feasible
solutions are ranked among them according to the objective of the MIP
problem, and infeasible solutions are ranked among them according to
the sum of infeasibilities (i.e., according to a measure of their
distance from the feasible set).  For this purpose, we define an
operator to compare two individuals.  We say that $\zeai \succ \zeaj$
($i$ \emph{is better} than $j$) iff:
\begin{itemize}
\item $i$ is feasible and $j$ is infeasible;
\item $i$ and $j$ are feasible, and $z^i > z^j$ ($i$ has a better
  objective);
\item $i$ and $j$ are infeasible, and $\zeta^i < \zeta^j$
  ($i$ is closer to the feasible region than~$j$).
\end{itemize}


As there is the possibility that both feasible and infeasible
individuals coexist in the population, their fitness cannot be
attributed as in common EAs, based only on the value of an objective
function.  Therefore, selection of an individual has to be (directly
or indirectly) based on its ranking in the population, which can be
determined through the comparison operator defined above (see also
section~\ref{sec:selection}).

\subsection{Initialisation}
\label{sec:initialisation}

The population that it used at the beginning an evolutionary process
is usually determined randomly, in such a way that the initial
diversity is very large.  In the case of MIP, it is appealing to bias
the initial solutions, so that they are distributed in regions of the
search space that are likely to be more interesting.  A way to provide
this bias, inspired in an algorithm provided in~\cite{gunluk96}, is to
firstly solve the LP relaxation of the problem, and then round the
solutions obtained to one of the closest integers.  The probabilities
for rounding up or down each of the variables are given by the
distance from the fractional solution to its closest integer points.

If we denote the solution of the LP relaxation by $\xlp = (\xlp_1
\ldots \xlp_n)$, each element of the initial population will be
determined as follows.  For all the chromosomes $k \in \{1, \ldots,
n\}$, the corresponding variable $\xea_k$ is rounded down with
probability
$$P(\xea_k = \lfloor \xlp_k \rfloor) = \xlp_k - \lfloor \xlp_k
\rfloor$$
or rounded up with probability $1 - P(\xea_k = \lfloor
\xlp_k \rfloor)$.

\subsection{The genetic operators}
\label{sec:gen-operators}

The generation of a new individual from two parents is composed of
three steps: recombination (meiosis and crossover), possibly followed
by mutation, followed by local search.  Each of the genetic operators
is controlled by two parameters: probability of occurrence and
intensity of the operation.

We use the following notation: $\nu^p$, $\chi^p$, $\mu^p$, are the
probabilities of meiosis, crossover, and mutation, respectively;
$\nu^s$, $\chi^s$, $\mu^s$ are their respective intensities.  The
distribution of the perturbations added by mutation is $\delta(s) = 1
- \rand^{s^2}$, were $s$ is the intensity and $\rand$ is a random
number uniformly distributed in $[0,1]$.  The value of $\delta(s)$ is
scaled, so that it covers the whole region between the value $\xea_k$
and its bounds.

The process of reproduction for creating a new genome $\xea$ from two
parents $\xeaf$ and $\xeam$ is presented in figure~\ref{fig:genetic}.

\begin{figure}[htbp!]
  \begin{center}
    
\begin{minipage}[tl]{115mm}
\begin{program}
\item[select parents ($\xeaf$, $\xeam$)] \comment{\ }
\item[procedure Reproduce($\xeaf$, $\xeam$)] \comment{\ }
\item[if $\rand<\nu^p$]  \comment{Do the meiosis with probability $\nu^p$}
  \begin{program}
  \item[for $k=1$ to $n$ do ] \comment{\ }
    \begin{program}
    \item[set $p := \rand (n-k+1) (1-\nu^s)$] \comment{Determine the
        size of the ``path'' to select from one of the parents
        (inversely proportional to the intensity of meiosis)}
    \item[if $\rand < 1/2$] \comment{With 50\% probability, copy from
        the father ($\xeaf$)}
      \begin{program}
      \item[for $l=1$ to $p$ do] \comment{\ }
        \begin{program}
        \item[if $\rand<\chi^p$] \comment{With some probability do crossover,}
          \begin{program}
          \item[set  $\xea_k := \xeaf_k + (\xeam_k - \xeaf_k)\; \chi^s \rand$]
            \comment{with intensity $\chi^s$}
          \end{program}
        \item[else] \comment{\ }
          \begin{program}
          \item[set  $\xea_k := \xeaf_k$] \comment{No crossover, exact copy of
              $\xeaf_k$} 
          \end{program}
        \end{program}
      \item[done] \comment{\ }
      \end{program}
    \item[else] \comment{With 50\% probability, copy from
        the mother ($\xeam$)}
      \begin{program}
      \item[. . .] \comment{(swap the roles of $\xeaf$ and $\xeam$)}
      \end{program}
    \item[end if] \comment{\ } 
    \end{program}
  \item[done] \comment{\ }
  \end{program}
\item[else ] \comment{In this case, no meiosis occurs:}
  \begin{program}
  \item[set $\xea := \xeaf$, or $\xea := \xeam$] \comment{copy exactly
      $\xeaf$ or $\xeam$, with same probability}
  \end{program}
\item[end if] \comment{\ }
\item[for $k=1$ to $n$ do ] \comment{Now, do the mutation: for each
    element of $\xea$,} 
  \begin{program}
  \item[if $\rand < \mu^p$] \comment{with probability $\mu^p$, add
      mutation}
    \begin{program}
    \item[set $\xea_k := $round($\xea_k \pm \delta(\mu^s)$)]
      \comment{of intensity $\mu^s$, and round to
        nearest integer.}
    \end{program}
  \end{program}
\item[done] \comment{\ }
\item[end procedure] \comment{\ }
\end{program}
\end{minipage}
    \caption{Pseudo programming code for the genetic operations.}
    \label{fig:genetic}
  \end{center}
\end{figure}

The recombination process produces a linear combination of the genomes
of two individuals selected from the population, and is based on two
sub-operations: meiosis and crossover.  Given two progenitor genomes,
the meiosis consists of selecting "paths", or sequences of $\xea_k$'s,
alternately from each of them, to create a new genome.  The greater
the meiosis intensity ($\nu^s$), the smaller these paths are likely to
be.  Crossover consists of, for each of the chromosomes (indices of
the genome vector), perturbing the value obtained by meiosis in the
direction of its value for the other progenitor. The smaller the
\emph{crossover intensity} parameter is, the closer the produced
chromosome is to that of one of the parents.

The mutation adds a random perturbation to the genome created this
way.  For each mutation we randomly choose, with identical
probability, to add or subtract $\delta(s)$ to the value of the
chromosome, where $s$ is the intensity, or magnitude of the mutation.
We then round the value to the closest integer.

Local search tries to improve the newly created individual's
performance by hill climbing in its neighbourhood, as described below.

\subsection{Local search}
\label{sec:local-search}

We propose a rather rough---but general---local search method, for
hill climbing in the integer variables space.  This search is
performed whenever a new individual is created.  It is based on what
is called \emph{hunt search}, originally conceived for locating values
in an ordered table.  The idea is to check for improvements in the
objective when each of the $n$ integer variables $\xea_k$ is
independently perturbed, with a geometrically increasing step.


The algorithm is the presented in figure~\ref{fig:local-search}.  Note
that this local search method is completely problem-independent, and
that its use does not exclude the possibility of using an additional,
problem-specific local search method to speedup the search.
\begin{figure}[htbp!]
  \begin{center}
\begin{minipage}[t]{115mm}
\begin{program}
\item[procedure Local\_Search($\xea$)] \comment{\ }
\item[set $\xls := \xea$] \comment{Start the local search at the
    individual's solution}
\item[for $k=1$ to $n$, do] \comment{For all the integer variables (in a
    random order):}
  \begin{program}
  \item[set $\xls_k := \xea_k+1$, $\zu :=$ LPsolution($\xls$)]
    \comment{Evaluate the perturbed solution}
  \item[set $\xls_k := \xea_k-1$, $\zd :=$ LPsolution($\xls$)]
    \comment{by solving the corresponding LP}
  \item[if $\zea \succ \zu$ and  $\zea \succ \zd$] \comment{Solution
      is a local optimum with respect to index $k$} 
    \begin{program}
    \item[continue with next $k$] \comment{\ }
    \end{program}
  \item[if $\zu \succ \zd$] \comment{Stepping up is better than
      stepping down}
    \begin{program}
    \item[set ${\mathit{step}} := 1$] \comment{\ }
    \end{program}
  \item[else] \comment{\ }
    \begin{program}
    \item[set ${\mathit{step}} := -1$] \comment{\ }
    \end{program}
  \item[while improving $\zea$ do] \comment{\ }
    \begin{program}
    \item[set $\xls_k := \xea_k+{\mathit{step}}$, $\zls :=$
      LPsolution($\xls$)] \comment{\ } 
    \item[if $\zls \succ \zea$] \comment{An improvement was
        found:}
      \begin{program}
      \item[set $\xea_k := \xls_k$, $\zea := \zls$] \comment{update
          the solution}
      \end{program}
    \item[set ${\mathit{step}} := 2 \times {\mathit{step}}$]
      \comment{Geometrically increase the step}
    \end{program}
  \item[done] \comment{\ } 
  \end{program}
  \item[done] \comment{\ }
\item[end procedure]
\end{program}
\end{minipage}
    \caption{The local search procedure.}
    \label{fig:local-search}
  \end{center}
\end{figure}

\section{Niche search}
\label{sec:niche-search}
Niche search is an evolutionary algorithm where the total population
is grouped into separate niches, each of which evolves independently
of the others for some (sub-)generations.  The claim is that this way,
as the global evolutionary search pursues, more localised searches are
done inside each of the niches.  The algorithm is therefore expected
to keep a good compromise between intensification of the search and
diversification of the population.  This method has some similarities
with that described in~\cite{muhlenbein94}, where \emph{competing
  subpopulations} play a role similar to that of the niches.  An
application of niche search to a specific combinatorial optimisation
problem has been shown in~\cite{pedroso98icec}; here, it is extended
to the general MIP case.

Niches are subject to competition between them.  The bad niches (i.e.,
those which have worse populations) tend to extinguish: they are
replaced by new ones, which are formed by elements selected from a
``good'' niche and the extinguishing one.  All the parameters that
control the genetic operators described in
section~\ref{sec:gen-operators} (mutation intensity and probability,
etc.), together with a selectivity factor, are assigned exogenously
and randomly to each newly created niche.  (The selectivity determines
how good an individual must be in relation to the average of the niche
in order to have a favoured probability of being selected for
reproducing.)

\subsection{Niche search core algorithm}
\label{sec:algorithm}

We summarise now the main steps of functioning for the niche search
algorithm.  This is the kernel algorithm, which drives the population
operations making use of the solution representation and genetic
operators described in the preceding sections.  Niche search is
characterised by evolution in two layers: in the higher layer, there
is the evolution of niches, subject to competition between them.  Each
iteration of this process is called a \emph{niche generation}, or
simply a generation.  In the lower layer, the individuals that
compose each niche evolve inside it, competing with other individuals
of the niche.  Each iteration of this lower layer process is called an
\emph{individual's generation}, or a subgeneration.

The code describing the evolution of the set of niches, in what we
call a niche generation, is presented in figure~\ref{fig:niche-search}.

\begin{figure}[htbp!]
  \begin{center}
\begin{minipage}[tl]{115mm}
\begin{program}
\item[set t := 0]  \comment{Start with an initial time.}
\item[niches(t) = CreateNiches(t)] \comment{Create desired number
    of niches for the run.}
\item[InitParameters(niches(t))] \comment{Randomly
    initialise the parameters that characterise each niche:
    crossover probability and intensity, mutation probability and
    intensity, etc.}
\item[InitialisePopulation(niches(t))] \comment{Randomly
    initialise the pop.\ of each niche.}
\item[Evaluate(niches(t))] \comment{Evaluate the fitness of all the
    niches in the initial population.  For evaluating a niche, we
    used the fitness of its best element (other strategies
    are also possible).}
\item[iterate]   \comment{Start evolution.}
\begin{program}
\item[Breed(niches(t))] \comment{Create a new generation of
    individuals in each of the niches, through the lower layer
    evolution process described below.}
\item[Evaluate(niches(t))] \comment{Evaluate the new niches.}
\item[set weak(t) := SelectWeak(niches(t))]     \comment{Select
    niches that will extinguish.}
\item[set strong(t) := SelectStrong(niches(t))]     \comment{Select
    niches that will be used for generating new niches.}
\item[set newniches(t) := Recombination(weak(t),strong(t))]
  \comment{Create a new niche for replacing each of the
    extinguishing ones.  The recombination strategy used is to
    create a population formed of the union of the weak niche with
    a strong one.  Then, replace the individuals of the weak niche
    by a selection of the best individuals from that population.}
\item[InitParameters(newniches(t))] \comment{Assign random
    parameters to new niches.}
\item[Evaluate(newniches(t))]     \comment{Evaluate the new niches.}
\item[Extinguish(weak(t), niches(t))] \comment{Remove weak
    niches from the population}
\item[Insert(newniches(t), niches(t))] \comment{and include the
    newly created ones.}
\item[set niches(t+1) := niches(t)] \comment{ }
\item[set t := t + 1]      \comment{Increase the time counter.}
\end{program}
\item[until Terminated()] \comment{Termination criteria: number of
    generations completed.}
\item[display solution] \comment{Solution is the best individual
    found.}
\end{program}
\end{minipage}
    \caption{Niche search: evolution of niches.}
    \label{fig:niche-search}
  \end{center}
\end{figure}

We now turn to the evolution of the individuals inside each of the
niches.  Pseudo-programming code describing how individuals breed at
each generation of the inside-niche evolution (i.e., describing what a
\emph{subgeneration} is) is presented in
figure~\ref{fig:inside-niche}.  Note that this process is repeated
for each of the niches, at each niche generation.

\begin{figure}[htbp!]
  \begin{center}
    \begin{minipage}[tl]{115mm}
      \begin{program}
      \item[Procedure Breed(niches(t))] \comment{ }
      \item[for all niche in niches(t) do] \comment{(\textup{\textsf{t}}
          is the \emph{niche generation} counter).}
        \begin{program}
        \item[set g := 0]  \comment{Initialise the ``subgeneration''
            counter.}
        \item[set population(g) := niche] \comment{Set the reference
            population: (only) the elements of the niche that is now
            breeding.}
        \item[iterate] \comment{Start evolution.}
          \begin{program}
          \item[for all element in offspring(g) do] \comment{ }
            \begin{program}
            \item[$p_1$ = Selection(population(g))] \comment{Select
                parents for reproduction (in our imple-} 
            \item[$p_2$ = Selection(population(g))]
              \comment{mentation through roulette wheel
                selection).} 
            \item[set element := Reproduce($p_1, p_2$)] \comment{Create
                the offspring using the}
            \end{program}
          \item[done] \comment{ operators described in
              (section\protect~\ref{sec:local-search}).}
          \item[Evaluate(offspring(g))] \comment{Evaluate the objective of
              all the individuals in the niche's population.  Scale to
              obtain the fitnesses (section\protect~\ref{sec:selection}).}
          \item[set population(g+1) := offspring(g)]
            \comment{Future population is the offspring.}
          \item[set g := g + 1]      \comment{Increase the
              subgeneration counter.}
          \end{program}
        \item[until Terminated()] \comment{Termination criteria: best
            individual has not improved.}
        \item[set niche := population(g)]
          \comment{Update niche's population.  This niche is now ready
            to start}
        \end{program}
      \item[done] \comment{competition with the others.}
      \item[end procedure] \comment{ }
      \end{program}
    \end{minipage}
    \caption{Niche search: evolution inside the niches.}
    \label{fig:inside-niche}
  \end{center}
\end{figure}

\subsection{Selection in each niche: rank-based fitnesses}
\label{sec:selection}

As explained in section~\ref{sec:objective}, the solution process is
divided into two goals: obtaining feasibility and optimisation.  This
has motivated the implementation of an order-based fitness attribution
scheme.  The selection of the individuals that are able to reproduce
at each generation is based on a fitness value, called
\textit{rank-fitness}, that is proportional to their ranking
according to the comparison operator defined in
section~\ref{sec:comparison}.

In niche search there is a parameter of each niche, called the
\textit{selectivity}, that controls the probability of selection of
each individual in relation to their competitors.  If this parameter
is very low, then the probability of selection of the best individuals
is only slightly greater than the probability of selection of the
worst; if it is high, then the best individuals have a much greater
probability of selection, what means that the ``genetic information''
of the worse ones is not likely to propagate to the future
generations.

In a niche with $n$ elements, the best of them is assigned a
rank-fitness of $1$ (i.e., $n/n$), the second-best $(n-1)/n$, up to
the worse, whose rank-fitness is $1/n$.  We then elevate this value to
a power, greater or equal to zero---the selectivity parameter of the
niche---to obtain the scaled-fitness of each individual.  The
selection is then performed through roulette wheel selection, giving
to each individual a probability of selection proportional to its
scaled-fitness (see, for example,~\cite{goldberg89} for a description
of roulette wheel selection).

\subsection{Elitism}
\label{sec:elitism}

Elitism determines whether the best solution found so far by the
algorithm is kept in the population or not.  Elitism generally
intensifies the search in the region of the best solution.  As
mentioned before, niche search keeps several groups, or niches,
evolving with some independence.  Each of these groups may be elitist
(keeping \emph{its} best element in its population) or not.

Our objectives are two fold: we want the search to be as deep as
possible around good regions, but we do not want to neglect other
possible regions.  The strategy that we devised for accomplishing this
is the following.  Niches whose best individual is different of the
best individual of other niches are elitist, but when several niches
have an identical best individual (and this occurs frequently), only
one of them is elitist.  With this strategy we hope to have an
intensified search on regions with good solutions, and at the same
time enforce a good degree of diversification.

\section{Numerical results}
\label{sec:results}

The instances of MIP problems used as benchmarks are defined in the
MIPLIB~\cite{miplib}.  The evolutionary system starts by reading an
MPS file, and stores the information contained there into an internal
representation.  The number of variables and constraints, their type
and bounds, and all the matrix information is, hence, determined at
runtime.

Note that the LPs solved by the EA are often much simpler than those
solved by B\&B; as all the integer variables are fixed, its size may
be much smaller (for a large proportion of integer variables).
Therefore, it is not surprising that numerical problems that the LP
solver may show up in B\&B, generally do not arise for LPs formulated
by the EA.

\subsection{Branch-and-bound}
In our implementation we have used a publicly available LP solver
called \emph{lp\_solve}~\cite{lp-solve} for the solution of the linear
programs.  This solver also comprises an implementation of the B\&B
algorithm, that was used for producing results to compare with the
evolutionary algorithm.

The B\&B scheme consists on depth-first search, branching on the first
non-integer variable.  Results obtained using B\&B on the series of
benchmark problems selected are provided in table~\ref{tab:bb-sols}.
The maximum number of LPs solved in B\&B was limited to 50 million; in
cases where this was exceeded, the best solution found within that
limit is reported.

\begin{table}[htbp!]
  \begin{center}
    \leavevmode
    \begin{tabular}[t]{|l|c|c|l|}
      \hline
      Problem   & Best solution & Number of     & Remarks \\
      name      & found         & LPs solved    & \\
      \hline \hline
      bell3a    &  878430.316 &     438737 & Optimal  \\
      bell5     & 8966406.492 &    2159885 & Optimal  \\
      egout     &      562.27 &      55057 & Solution incorrect (rounding problems?) \\
      enigma    &           0 &       9321 & Optimal  \\
      flugpl    &     1201500 &       2179 & Optimal  \\
      gt2       &         --  &         -- & Failed (unknown error)         \\
      lseu      &        1120 &     252075 & Optimal  \\
      mod008    &         307 &    2848139 & Optimal  \\
      modglob   & 27124594.43 &$>$50000000 & Stopped (excessive CPU time) \\
      noswot    & -23.0 (infeas.) &   3042 & Failed (numerical instability?)\\
      p0033     &        3089 &       7409 & Optimal  \\
      pk1       &          12 &     704208 & Failed (numerical instability)\\
      pp08a     &        9880 &$>$50000000 & Stopped (excessive CPU time) \\
      pp08acut  &        7900 &$>$50000000 & Stopped (excessive CPU time) \\
      rgn       &        82.2 &       4747 & Optimal  \\
      stein27   &          18 &      12031 & Optimal  \\
      stein45   &          30 &     235087 & Optimal  \\
      vpm1      &          21 &    1685443 & Failed (numerical instability?)\\
      \hline
    \end{tabular}
    \caption{Solutions obtained for branch-and-bound.}
    \label{tab:bb-sols}
  \end{center}
\end{table}

\subsection{Evolutionary algorithm}

Niche search was used to make 5 niches, each with 5 individuals,
evolve for 250 niche generations.  In each of these generations, the
population of each niche would reproduce until no improvements in its
best element were observed.  Although tuning up the population and
generation numbers would likely lead to better results, we have made
not attempt to do so, and used the same values for all the problems.

The MIPLIB minimisation problems were converted into maximisations.

In table~\ref{tab:ea-sols} we report the optimal solutions, as stated
in the MIPLIB, and the range of the final solutions determined in an
experiment with 25 independent runs of niche search for each of the
benchmark problems.  For more than 50\% of the tests, the optimal
solution could be determined.  The EA failed to systematically find a
feasible solution only for the \emph{enigma} problem.  
The average number of LPs that were solved until obtaining the
solutions for niche search reported is written on the rightmost
column.

\begin{table}[htbp!]
  \begin{center}
    \leavevmode
    \begin{tabular}[t]{|l|r||r|r|r||r|}
      \hline
\mcc{Problem} & \mccr{Optimal} & \multicolumn{4}{|c|}{Niche search
  solutions (25 runs)} \\ \cline{3-6}
\mcc{name}   & \mccr{solution} & \mcc{Worst} & \mcc{Mean} &
\mccr{Best}& Avg.\#LPs \\ \hline\hline
bell3a   & -878430.32 & -1502340 & -929125.2  & -881935  & 312455 \\
bell5    & -8966406.49& -9342570 & -9121121.2 & -9030450 & 597709 \\
egout    & -568.101   & -575.983 & -568.73156 & -568.101 & 48843\\
enigma   & -0.0       & -24 \small{\textit{(infeas.)}}& 
                        -15.2 \small{\textit{(infeas.)}}& 
                        -7\small{ \textit{(infeas.)}} & 252582 \\
flugpl   & -1201500   & -1240500 & -1209300   & -1201500 & 37478\\
gt2      & -21166.0   & -42006   & -32375.84  & -22342   & 1249782\\
lseu     & -1120      & -1542    & -1270.12   & -1120    & 232348 \\
mod008   & -307       & -349     & -317.04    & -307     & 154303\\
modglob  & -20740508  & -20740508&  -20740508 &-20740508 & 99478\\
noswot   &  +43       & 29       & 39.56      & 41       & 401825 \\
p0033    & -3089      & -3188    & -3097.64   & -3089    & 25785\\
pk1      & -11        & -29      & -23.6      & -19      & 93031  \\
pp08a    & -7350      & -7390    & -7358.8    & -7350    & 45780\\
pp08acut & -7350      & -7350    & -7350      & -7350 & 45582   \\
rgn      & -82.1999   & -82.2        &  -82.2       &  -82.2     & 8050\\
stein27  & -18        & -18      &  -18       &  -18     & 286  \\
stein45  & -30        & -31      & -30.04     & -30      & 43108\\
vpm1     & -20        & -20      &  -20       &  -20     & 7397 \\
\hline
    \end{tabular}
    \caption{Optimal solutions of the benchmark problems reported in
      MIPLIB, solutions obtained in an experiment with 25 independent
      runs of niche search, and average number of LPs solved for
      obtaining them.}
    \label{tab:ea-sols}
  \end{center}
\end{table}

\begin{table}[hbtp!]
  \begin{center}
    \leavevmode
    \begin{tabular}[t]{|l||r|r||r|r||c|}
      \hline
Problem name & $r^f/R$  & $E[n^f]$   &  $r^o/R$   & $E[n^o]$ & Best algorithm \\
\hline \hline                                 
bell3a   &   100\%    & 2053       &    0\% &$>$18246645 & B\&B\\
bell5    &   100\%    & 33748      &    0\% &$>$18024642 & B\&B\\
egout    &   100\%    &  423       &   92\% & 133764     & EA  \\
enigma   &     0\%    & $>$11876637&    0\% &$>>$11876637& B\&B\\
flugpl   &   100\%    &  29048     &   80\% & 91004      & B\&B\\
gt2      &   100\%    & 6383       &    0\% &$>$37665907 & EA  \\
lseu     &   100\%    & 1985       &    4\% &  10269416  & B\&B\\
mod008   &   100\%    &  17        &   48\% & 2557585    & EA? \\
modglob  &   100\%    &  3         &  100\% &      99478 & EA  \\
noswot   &   100\%    & 33627      &    0\% &$>$34335094 & EA  \\
p0033    &   100\%    &  8350      &   80\% & 93571      & B\&B\\
pk1      &   100\%    & 3          &    0\% &$>$6259152  & EA  \\
pp08a    &   100\%    &  49        &   72\% & 177969     & EA  \\
pp08acut &   100\%    &  33      &   100\% & 45582       & EA  \\
rgn      &   100\%    &  21        &  100\% & 8050       & B\&B\\
stein27  &   100\%    &  41        &  100\% &      286   & EA  \\
stein45  &   100\%    &  61        &   96\% & 54791      & EA  \\
vpm1     &   100\%    &  123       &  100\% &      7397  & EA  \\
\hline
    \end{tabular}
    \caption{Niche search: number of successes and expected
      number of LP solutions for finding a feasible and the optimal
      solution, respectively, and performance comparison with B\&B.}
    \label{tab:ea-summary}
  \end{center}
\vspace{-10mm}  
\end{table}

In order to assess the empirical efficiency of the algorithm, we
provide a measure of the expectation of the number of LP solutions
required for finding a feasible and the optimal solution.  Let $R$ be
the number of runs per benchmark problem in a given experiment, and
$r^f$ and $r^o$ be the number of runs in which a feasible and the
optimal solution are found, respectively.  Let $n^f_i$ be the number
of LP solutions that were required for obtaining a feasible solution
in run $i$, or the total number of LPs solved in that run if no
feasible solution was found.  Similarly, let $n^o_i$ be the number of
LP solutions required for reaching optimality, or the total number of
LPs solved in $i$ if no optimal solution was found.  Then, the
expected number of LPs for reaching feasibility, based on these $R$
observations, is:
$$E[n^f] = \sum_{i=1}^{R} \frac{n^f_i}{r^f}$$
Equivalently, the expected number of LPs for reaching optimality is
$$E[n^o] = \sum_{i=1}^{R} \frac{n^o_i}{r^o}$$
These values are
reported for each of the benchmark problems in
table~\ref{tab:ea-summary}.  On the case of $r^o=0$, the sum of the LP
solutions of the total experiment ($R$ runs) provides a lower bound on
the expectations for optimality.  The same for feasibility, when
$r^f=0$.  These are the values reported in table~\ref{tab:ea-summary}
for those situations.  In this table we also make a comparison of B\&B
and the EA.  The judgement is based on the reliability and on the
expected number of LPs required for optimality, for each of the
algorithms.
      
For some problems (e.g. pp08a) the EA quickly obtained a good
solution, even though B\&B has failed.  For other (e.g.\ modglob,
mod008), a feasible solution was easily found at the beginning of the
EA, suggesting its possible use as a method for obtaining a feasible
solution to speedup B\&B.  Some benchmarks---especially
\textit{enigma}---were easily solved by B\&B, even though the EA had
problems tackling them.

\begin{figure}[htbp!]
  \begin{center}
    \epsfig{file=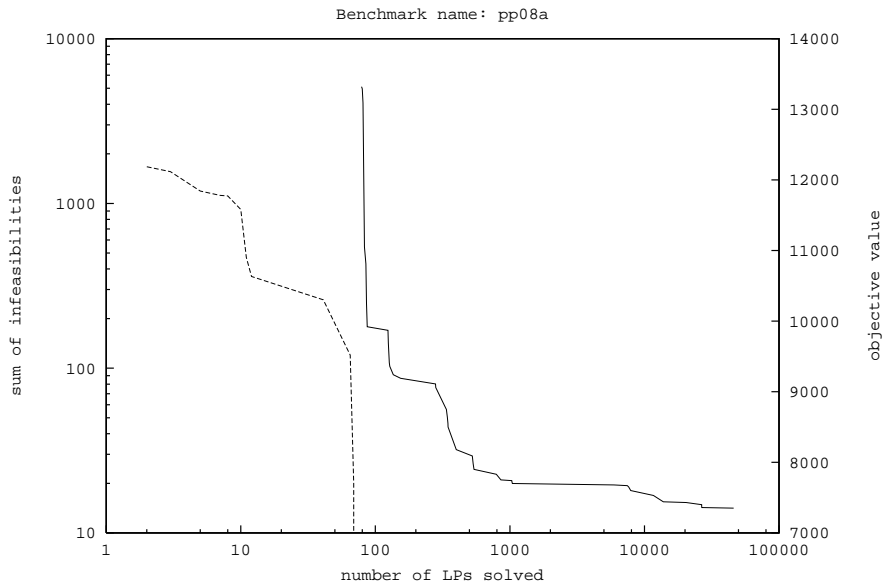,width=80mm,angle=-90}
    \caption{Typical log of the evolution of the solution with the
      number of LPs solved, in this case for the \emph{pp08a}
      benchmark.  A feasible solution was found at around the 70th
      LP solved.  Dotted line for infeasible solutions (left side $y$
      axis), continuous line for feasible ones (right side $y$ axis).}
    \label{fig:graph}
  \end{center}
\end{figure}

In figure~\ref{fig:graph} is plotted a log of the evolution of the
population's best solution in a typical run of the EA, for the case of
the \emph{pp08a} problem.  The curve at the beginning of the process
corresponds to infeasible solutions; a feasible solution is found in
the middle of the process.  In most of the cases, these two phases of
the search can be distinctly observed: first minimising the
infeasibilities and then, when a feasible solution is found,
optimising the objective.

In order to assert the importance of each of the operators used in the
evolutionary system, we executed some experiments for assessing their
efficiency.  These experiments consisted on keeping track of which of
the operators were responsible for improvements on the solutions, and
of analysing the behaviour of the algorithm in their absence.  They
showed that the three genetic operators, the local search, and the
initialisation procedure, where all necessary for a good performance
of the algorithm.

We also made a series of runs with only one niche, increasing the
number of generations so that the maximum number of LP solutions was
approximately the same as the one used for the results reported in
this section.  The solutions obtained provided an empirical
confirmation of the importance of the separation of the population in
niches.  With a single niche the algorithm decreased its performance,
both in terms of the number of runs that lead to feasibility and
optimality, and in terms of the number of calls to the LP solver that
were required for obtaining an equivalent final solution.

\section{Conclusion}
\label{sec:conclusion}

In this paper we present an evolutionary algorithm for the solution of
integer linear programs based on the separation of the variables into
the integer part and the continuous part.  The integer variables are
fixed by the evolutionary system, and replaced in the original LP,
leading to a pure continuous problem.  The optimisation of this LP
determines the continuous variables corresponding to that integer
solution, and the objective value leads to the solution's fitness.

The results obtained for some of the standard benchmark problems were
compared to those obtained by B\&B.  The performance of the
evolutionary algorithm is promising.  In some of the benchmark tests
it outperformed B\&B, either by requiring less LP solutions to
systematically reach the optimum, or by succeeding in determining a
good feasible---sometimes optimal---solution in cases where B\&B
failed.

The success of the algorithm in finding good feasible solutions with
limited computational resources for most of the benchmark problems
testify its potentialities for real-world, practical applications.

The algorithm proposed does not take into account any particular
structure of the problems (it is based only on the information
contained in MPS files; nothing about the specific kind of problem
dealt with is taken into account).  For obtaining more competitive
results, a problem-specific local search, exploiting the particular
structure of the problem, should be additionally implemented.

The discrepancy between the results obtained by the EA and by B\&B
suggests that these algorithms are probably good complements of each
other, and the integration of both approaches in a single tool seems
to be a promising research direction.

An advantage, not yet exploited, of this evolutionary algorithm is
that the models that it can tackle may include non-linearities, as
long as a linear problem can be obtained by fixing some variables.
These nonlinear variables would also be fixed by the evolutionary
structure, at the time of fixing the integer ones, in such a way that
the resulting problem is linear and continuous.

\section{Acknowledgements}
This work was supported by the European Union Science and Technology
Fellowship Programme in Japan.

\appendix
\section{Appendix}

\subsection{Implementation details}
With the aim of warranting the possibility of reproduction of the
results presented in this paper, we provide some details on the
implementation of the niche search algorithm.


\subsubsection{Avoiding re-evaluation}
Sometimes the genetic operations do not lead to a different
individual, the solution generated is identical to one of the parents.
In this case, the newly generated individual will also carry the LP
solution information from the parent, and is not re-evaluated.  Local
search was already done on the neighbourhood of the solution
corresponding to the identical parent, and hence it is not performed
again.

\subsubsection{Niche recombination}
Whenever a niche is ``extinct'', a new one, with new parameters, is
created (see figure~\ref{fig:niche-search}), in what we call
\textit{niche recombination}.  This process starts by selecting the
best niche, subject to the restriction that its best element is not
present in other niches.  We then make the union of this niche's
population with the population of the extinguishing one, and select
the best distinct elements from this pool into the new niche.  If the
pool is not diverse enough, the number of distinct elements is less
than the population of the new niche.  In this case, the remaining
individuals are initialised as described in
section~\ref{sec:initialisation}.

\subsubsection{Parameters used}
At each generation the number of niches that may extinguish is 35\% of
the total.  The probability of extinction is 35\%.  All the other
parameters (the probability and intensity of meiosis, crossover and
mutation, and the selectivity) are different for each niche.  They are
assigned randomly, with uniform distribution, whenever a new niche is
created: at the begin of the evolution, or in a niche recombination.

\nocite{goldberg89}
\nocite{pedroso96ppsn}
\bibliography{../../jpp} \bibliographystyle{plain}

\end{document}